\documentclass[conference]{IEEEtran}
\IEEEoverridecommandlockouts
\usepackage{cite}
\usepackage{amsmath,amssymb,amsfonts}
\usepackage{algorithmic}
\usepackage{graphicx}
\usepackage{textcomp}
\usepackage{xcolor}
\usepackage{hyperref}
\def\BibTeX{{\rm B\kern-.05em{\sc i\kern-.025em b}\kern-.08em
    T\kern-.1667em\lower.7ex\hbox{E}\kern-.125emX}}
\makeatletter 
\newcommand{\linebreakand}{%
  \end{@IEEEauthorhalign}
  \hfill\mbox{}\par
  \mbox{}\hfill\begin{@IEEEauthorhalign}
}
\makeatother 
\usepackage[pscoord]{eso-pic}
\begin{document}

\newcommand{\placetextbox}[3]{
\setbox0=\hbox{#3}
\AddToShipoutPictureFG*{ \put(\LenToUnit{#1\paperwidth},\LenToUnit{#2\paperheight}){\vtop{{\null}\makebox[0pt][c]{#3}}}
}
}
\title{Experimental Study on Automatically Assembling Custom Catering Packages With a 3-DOF Delta Robot Using Deep Learning Methods
}
\author{\IEEEauthorblockN{Reihaneh Yourdkhani, Arash Tavoosian, $^*$Navid Asadi Khomami and Mehdi Tale Masouleh}\IEEEauthorblockA{Email: \{r.yourdkhani, arash.tavoosian, navid.asadi, m.t.masouleh\}@ut.ac.ir}\IEEEauthorblockA{Human and Robot Interaction Laboratory}\IEEEauthorblockA{Electrical and Computer Engineering Department, $^*$Mechanical Engineering Department}\IEEEauthorblockA{ University of Tehran}}
\makeatletter
\let\old@ps@headings\ps@headings
\let\old@ps@IEEEtitlepagestyle\ps@IEEEtitlepagestyle
\def\confheader#1{%
\def\ps@headings{
\old@ps@headings
\def\@oddhead{\strut\hfill#1\hfill\strut}
\def\@evenhead{\strut\hfill#1\hfill\strut}
}
\def\ps@IEEEtitlepagestyle{
\old@ps@IEEEtitlepagestyle
\def\@oddhead{\strut\hfill#1\hfill\strut}
\def\@evenhead{\strut\hfill#1\hfill\strut}
}
\ps@headings
}
\makeatother
\confheader{
\small{2024 32nd International Conference on Electrical Engineering    } 
}
\placetextbox{.23}{0.055}{\textbf{\small{979-8-3503-7634-0/24/\$31.00~\copyright 2024 IEEE}}}
\maketitle

\begin{abstract}

This paper introduces a pioneering experimental study on the automated packing of a catering package using a two-fingered gripper affixed to a 3-degree-of-freedom Delta parallel robot. A distinctive contribution lies in the application of a deep learning approach to tackle this challenge. A custom dataset, comprising 1,500 images, is meticulously curated for this endeavor, representing a noteworthy initiative as the first dataset focusing on Persian-manufactured products. The study employs the YOLOV5 model for object detection, followed by segmentation using the FastSAM model. Subsequently, rotation angle calculation is facilitated with segmentation masks, and a rotated rectangle encapsulating the object is generated. This rectangle forms the basis for calculating two grasp points using a novel geometrical approach involving eigenvectors. An extensive experimental study validates the proposed model, where all pertinent information is seamlessly transmitted to the 3-DOF Delta parallel robot. The proposed algorithm ensures real-time detection, calibration, and the fully autonomous packing process of a catering package, boasting an impressive over 80\% success rate in automatic grasping. This study marks a significant stride in advancing the capabilities of robotic systems for practical applications in packaging automation.\\

Index Terms- Catering Packages Dataset(CPO), Delta Parallel Robot(DPR), Grasping, Gripper
\end{abstract}
\begin{figure}[t!tbp]
\centerline
{\includegraphics[width=0.5\textwidth]{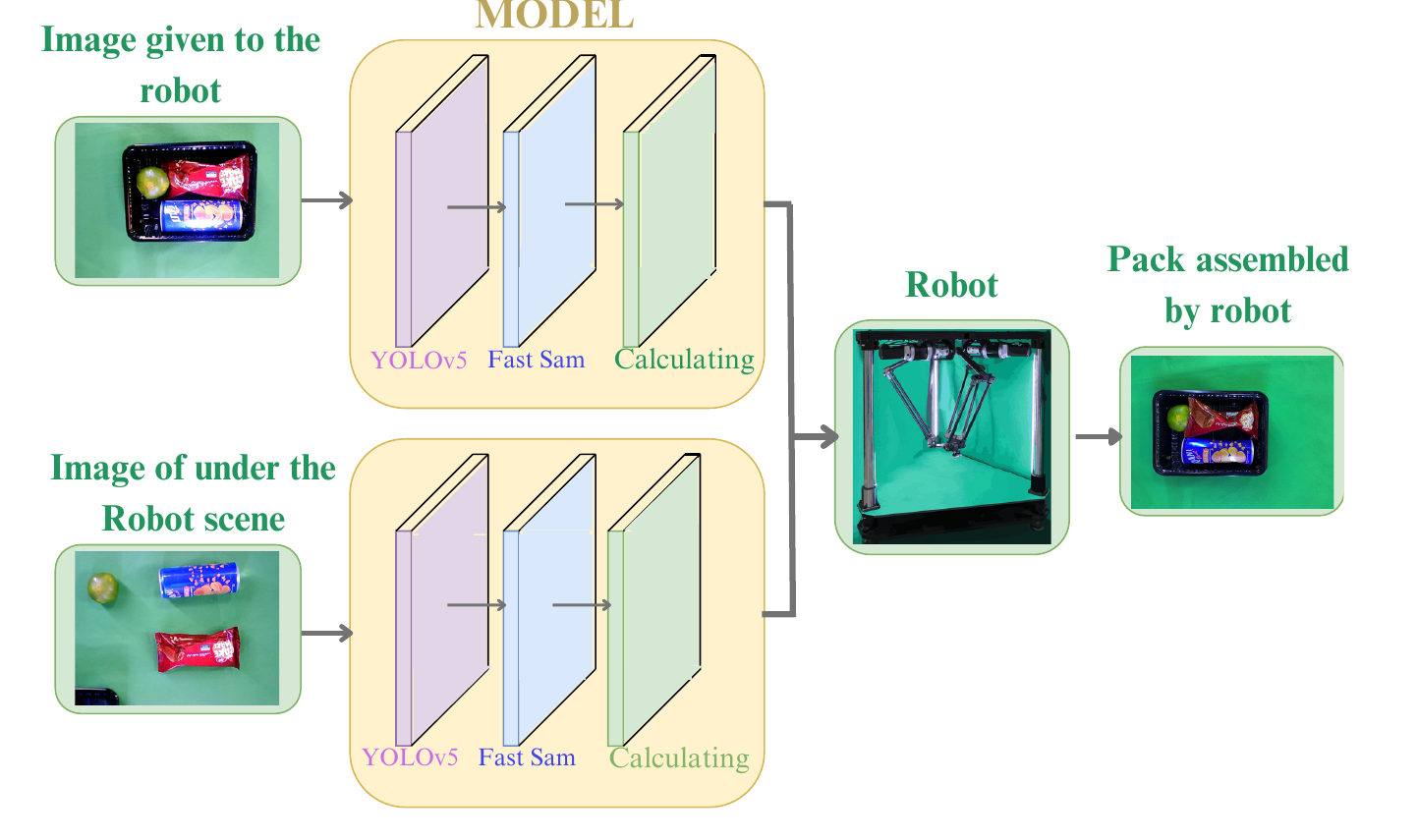}}
\caption{This figure demonstrates how two images get imported to the later proposed network, one is an image of the intended pack and the other objects under the robot. After detection, segmentation and geometrical calculations, the DPR and the gripper take action to pack the required objects according to the input image.}
\label{koll}
\end{figure}
\section{Introduction}
With 7.88 billion people on earth in 2023 and their increasing population, their demand for food has also increased. Global food demand is expected to increase by 35\% to 56\% between 2010 and 2050 \cite{van2021meta}. With that in mind, the COVID-19 pandemic brought much attention to hygiene in the food industry. According to scientists, cleaning, sanitation, good hygienic practices, and active packaging are needed from farm to fork \cite{olaimat2020food}. As a result, 1-increasing demand for food and 2-the importance of hygiene in the food industry made researchers and industries endeavor hand in hand to find new and innovative approaches to make robots facilitate this task in a sanitized environment faster than humans resorting to robotic mechanical systems. Therefore, the main objective of this paper consists of one of the most challenging autonomous food-related processes, i.e., packing catering packages. From packages served on a train or a bus to the ones in formal meetings, they all use mostly the same type of products: fruits, cans, juices, cakes and biscuits, coffee or tea, and disposables such as forks, spoons, and cups.

In the dynamic field of robotics, prior works form the bedrock of innovation. This study\cite{accorsi2019application}, emphasized the role of robots in food packaging, focusing on efficiency and precision. This paper\cite{liu2023food}, discusses the design and implementation of a fully automatic food package recognition and sorting system using machine vision and robotic arm technology to address challenges posed by reflective or transparent materials and the need for precise $z$-axis determination in grasping tasks. This paper\cite{ummadisingu2022cluttered}, presents a method for automating pick-and-place tasks in the food packaging industry by training instance segmentation models purely on synthetic data, transferring them to the real world using sim2real methods, and addressing concerns of food damage during grasping with adaptive finger mechanisms and grasp filtering techniques. In a continuation of the groundbreaking research conducted in the laboratory, the same robotic system has been further utilized to meticulously package chocolates with unparalleled precision. This innovative study not only underscores the versatility of the technology but also opens avenues for precise manipulation and handling in diverse industries \cite{mojtahedi2024experimental}. Yet, a noticeable gap exists in exploring intelligent models for personalized packaging\cite{roudbari2023autonomous}. This research goes beyond automation, utilizing intelligent models to discern and adapt to unique requirements for each custom package. By delving into intelligent robotic systems, the aim is to revolutionize food packaging, ushering in a more adaptive, responsive, and personalized process. The developed model transcends the limitations of specific industries, offering a groundbreaking solution for automating packaging across diverse sectors. Beyond food, it enables automated packing of various packages in any form, empowering industries with unprecedented autonomy and intelligence through customizable input pictures—a pioneering advancement previously unexplored in packaging automation.

The 3-Degree-Of-Freedom (DOF) Delta Parallel Robot (DPR) is one of the main robots used in the food industry for pick-and-place purposes and has found a prominent place in the latter market. This robot has the advantages of fast motion speed, accurate positioning, low cost and high efficiency. The latter features are the main reasons why the 3-DOF DPR with FDA-compliant components is suitable for gripping and depositing food.

Before initiating pick-and-place operations, object detection is paramount. Leveraging advancements in deep learning, particularly models like RCNN, FastRCNN, and FasterRCNN, has become instrumental\cite{xie2021oriented,girshick2015fast,ren2015faster}. For this study, deep learning methods are employed to integrate detection and segmentation models with conventional approaches. Creating a custom dataset tailored to the robot environment is a prerequisite for developing a specialized object detection model. After comparing YOLOv5 and YOLOv8, the YOLOv5 model with 100 epochs is chosen as the foundational layer. Preceding the grasping procedure, factors such as object orientation, coordination, and width are crucial. A combination of classic and modern approaches, including the efficient FastSAM object segmentation model, is employed to measure these factors accurately. This process yields precise information necessary for a successful grasp, subsequently communicated to the robot. Consequently, the robot, equipped with an end-effector gripper, autonomously performs packaging in a sterile environment, enhancing efficiency and speed without human intervention.

The key contributions of the proposed approach are the following:\begin{itemize}
\item Gathering a dataset on the items in a catering package, especially Persian manufactured objects, to the end of training the best network possible for detecting these objects.
\item Measuring required information for grasping with the help of the 3-DOF DPR and a gripper, using detection, segmentation and geometrical approaches.
\item Evaluating the performance of the proposed model for automatically assembling custom catering packages by performing practical test based on real scenario.  
\end{itemize}

The remainder of this paper is organized as follows. In Section \ref{sec:section2}, a comprehensive exploration of the methodology unfolds, delving into the dataset, the model, encountered challenges, and the DPR. Practical results will be discussed in Section \ref{sec:section6}. The final section encapsulates the conclusion and outlines ongoing works.
\section{Methodology}
\label{sec:section2}
In the following section, an overview of the methodology will be provided, encompassing the acquisition and analysis of the provided dataset, the description of the proposed model itself, and the discussion of the challenges encountered throughout the process.
\subsection{Catering package objects dataset}
\label{sec:section2-1}
 In order to gather comprehensive data for catering packages, a diverse set of objects crucial to these packages was considered. As these objects, mainly manufactured products, are influenced by brand characteristics like shape and color, a specific dataset, Catering Packages Objects (CPO), was created for this study. Notably, to the best knowledge of the authors of this paper, there were no existing datasets for Persian-manufactured products, necessitating the formation of the CPO dataset, which encompasses 1400 RGB images across 19 classes, and 4000 annotations in total. Tailored to the robotic environment, this dataset factors in conditions such as lighting and shadows. Additionally, two augmentation steps—rotations between -15 and 15 degrees and the introduction of noise—simulate real-world testing conditions, resulting in a comprehensive dataset of 3500 images. This dataset has been preserved and labeled using Roboflow platform for transparency and reproducibility\cite{roboflow}.

\subsubsection{Dataset Sources}\label{AA}
The CPO dataset is divided into two segments, enhancing its robustness. Approximately 90\% of the dataset comprises images captured by the authors of this paper, ensuring a controlled and consistent environment. The remaining 10\% is sourced from online shopping platforms through web scraping, introducing diversity and resilience to varied environments. This subset includes images obtained directly from manufacturers, characterized by their cleanliness and clarity, as well as images contributed by consumers in the comment sections—typically unclean and noisy, reflecting real-world scenarios. This dual sourcing strategy fortifies the dataset's adaptability and relevance across different contexts. The CPO dataset images exhibit two distinct groups based on capture devices: 1) iPhone camera images focusing on the test environment background, and 2) 720p webcam images taken under the robot, considering background, lighting (DPR LEDs and room lighting), and shadows. The latter set, pivotal for the experimental stage, introduces unique challenges explored in subsequent sections.

\subsubsection{Final results and analytics of the CPO dataset}
Given the challenges to be discussed later, a rationale for the uneven distribution of the CPO dataset could be justified. However, efforts have been made to circumvent this by recognizing the potential for object detection models to become overfitted to specific objects\cite{salman2019overfitting}. Through various tests, it was shown that the underrepresented data does not compromise the detection process. The dataset exhibits strength and variability, ensuring effective detection of any object within its scope. The CPO dataset at hand comprises 1400 images across 19 classes, with an average of 500 annotations per class. The minimum number of annotations is 300 for the "juice (standing)" class, while the maximum is 900 for the "logo" class.
\begin{figure}[ttbp]
\centerline{\includegraphics[width=0.47\textwidth]{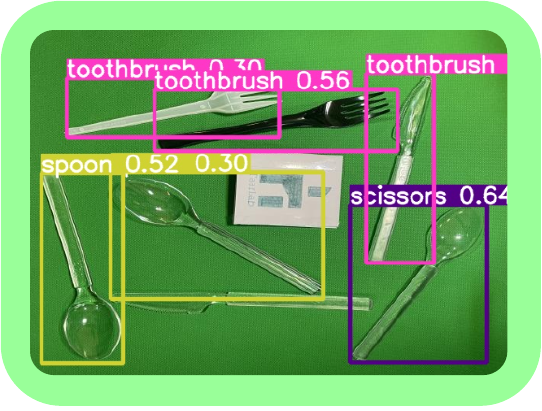}}
\caption{Wrong detections of YOLOV5 pre-tuned model.}
\label{yoloready}
\end{figure}
\subsection{Model}
\label{sec:section4}
Once the dataset is gathered accurately, the next crucial step is crafting a model. This model is geared towards efficiently detecting, segmenting, and conducting necessary calculations. The forthcoming subsection will delve into the details of this process.
\subsubsection{Object detection}
\label{sec:sectionod}
In the realm of deep learning, object detection stands out as a challenging and pivotal field which has garnered significant attention from researchers in recent years \cite{zhao2019object}. A prominent algorithm in this domain is YOLO (You Only Look Once), which, despite having various versions, shares key features \cite{hussain2023yolo}. Notably, these versions offer pre-tuned models capable of detecting specific objects, some of which align with the objects under consideration in this study—such as spoons, forks, knives, cups, and certain fruits.

In Fig.~\ref{yoloready}, the results of testing the pre-tuned YOLOv5 model on a sample image from the study's environment are displayed. Notably, the model's accuracy in detecting the aforementioned common objects is suboptimal, falling below 40\%. This limitation underscores the need for a dedicated dataset and the fine-tuning of a custom model. Consequently, in this study, YOLO has been fine-tuned using the CPO dataset, as detailed in the previous section.

In the realm of autonomous systems, YOLO's real-time object detection capabilities have garnered significant attention and found applications in diverse fields, including human-robot interaction \cite{jiang2022review}. Notably, among the various iterations of YOLO, both YOLOv5 and YOLOv8 have demonstrated commendable speed and accuracy in their results \cite{terven2023comprehensive}. Evaluating these models on the CPO dataset revealed similar precision, recall, and accuracy metrics. Considering YOLOv5's superior speed and lightweight architecture, it was selected for this study, trained over 100 epochs without altering hyperparameters.

The fine-tuned YOLOv5 model yielded impressive results: precision at 91\%, recall at 90\%, and a mean Average Precision at 50 (mAP50) of 93.2\%. This fine-tuning process optimized weights, notably decreasing loss functions, particularly class loss, across both training and validation datasets. Given that this step outputs object labels exclusively, a crucial metric for comparison is class loss, assessing the accuracy of each predicted bounding box's classification—whether it contains an object class or represents the background.

Metrics results on the test subset of the dataset are presented in Table \ref{prm}. Notably, objects like tangerines and bananas, with dissimilarities from the rest, exhibit exceptionally high metrics results. Some objects, such as logo, teabag, and Nescafe, display lower metrics results due to their shape similarities. However, as evidenced in Table \ref{grasp}, this discrepancy does not impede the real-time detection process and is primarily attributed to the web scraping aspect(noisy images) of the dataset rather than the captured images(clean real images).
\begin{table}[ttbp]
\caption{Each object's label vs. its evaluation metrics for detection on test dataset.}
\begin{center}
\resizebox{8cm}{!}{
\begin{tabular}{c|c|c|c}
\textbf{label}&\textbf{precision} & \textbf{\textit{recall}}& \textbf{\textit{mAP50}}\\ 
\hline
Banana&1&0.988&0.995 \\
\cline{1-4} 
Biscuit&0.917&0.906&0.977  \\
\cline{1-4} 
Cake&0.92&1&0.99  \\
\cline{1-4} 
Cup(laying)&0.824&0.781&0.852  \\
\cline{1-4} 
Cup(standing)&0.85&0.67&0.76\\
\cline{1-4} 
Fork&0.96&0.90&0.975\\
\cline{1-4} 
Juice(laying)&0.987&0.963&0.975  \\
\cline{1-4} 
Juice(standing)&1&0.97&0.995  \\
\cline{1-4} 
Knife&0.973&0.966&0.991\\
\cline{1-4} 
Logo&0.751&0.721&0.779  \\
\cline{1-4} 
Nescafe&0.87&0.779&0.862  \\
\cline{1-4} 
Nescafe(square)&0.925&0.95&0.905  \\
\cline{1-4} 
Pack&0.984&1&0.995  \\
\cline{1-4} 
Rani(laying)&0.878&1&0.935  \\
\cline{1-4} 
Rani(standing)&0.771&1&0.995  \\
\cline{1-4} 
Spoon&0.951&0.951&0.984  \\
\cline{1-4} 
Straw&0.951&0.951&0.984  \\
\cline{1-4} 
Tangerine&0.987&1&0.995  \\
\cline{1-4} 
Teabag&0.853&0.726&0.813  \\
\cline{1-4} 
Average&0.911&0.904&0.932  \\
\end{tabular}
}
\label{prm}
\end{center}
\end{table}

\subsubsection{ Segmentation}
After labeling each object, the next step is to get the important measurements. To do that, the objects need to be outlined or separated from the background.
\begin{figure}[ttbp]
\centerline{\includegraphics[width=0.5\textwidth]{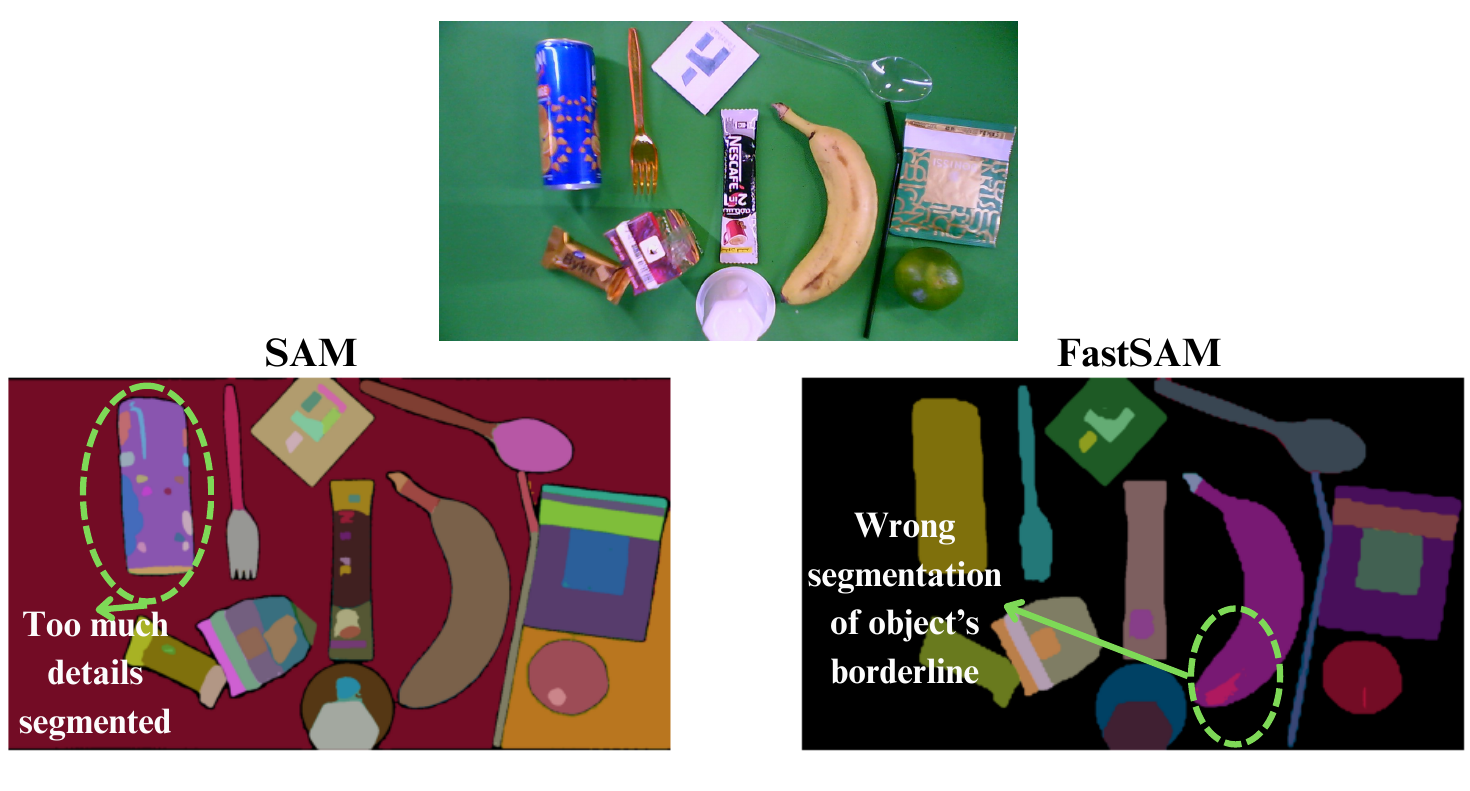}}
\caption{SAM vs FastSAM segmentation output using everything mode.}
\label{svsfs}
\end{figure}
In recent advancements, the Segment Anything Model (SAM) has emerged as a groundbreaking approach for object segmentation, requiring no prior training \cite{kirillov2023segment}. SAM utilizes user prompts, such as bounding boxes, text, points, or segment-everything inputs. Despite its versatility in various computer vision tasks, SAM's slow processing speed (2099 ms/image) raises concerns about its real-time applicability. To address this, a faster alternative, FastSAM, was introduced as a real-time solution for segmentation across diverse tasks \cite{zhao2023fast}. FastSAM boasts a remarkable 50-fold increase in speed compared to SAM, acknowledging the inherent tradeoff between speed and accuracy.

Figure~\ref{svsfs} provides a visual comparison of SAM and FastSAM outputs on a sample image from the dataset. SAM demonstrates higher accuracy in object detection, meticulously segmenting every intricate detail. In contrast, FastSAM, while 50 times faster, requires some corrections to accurately delineate object borders. Despite this, the study adopts an algorithm that selects the largest mask within a bounding box, mitigating the impact of minor discrepancies from FastSAM. The speed advantage of FastSAM in real-time segmentation outweighs its lower accuracy, facilitating its use in this study. Consequently, the bounding boxes generated by YOLOv5 in each scene serve as inputs to FastSAM, enhancing the precision of object segmentation.
\subsubsection{Geometrical methods for measuring the required parameters}
Upon reaching this stage, it becomes imperative to compute certain geometrical parameters before proceeding to the grasping phase. These calculations encompass determining the precise coordinates, orientation, and width of each object. These matters are elaborated in details in the remaining parts of this subsection.

\begin{figure}[ttbp]
\centerline{\includegraphics[width=0.5\textwidth]{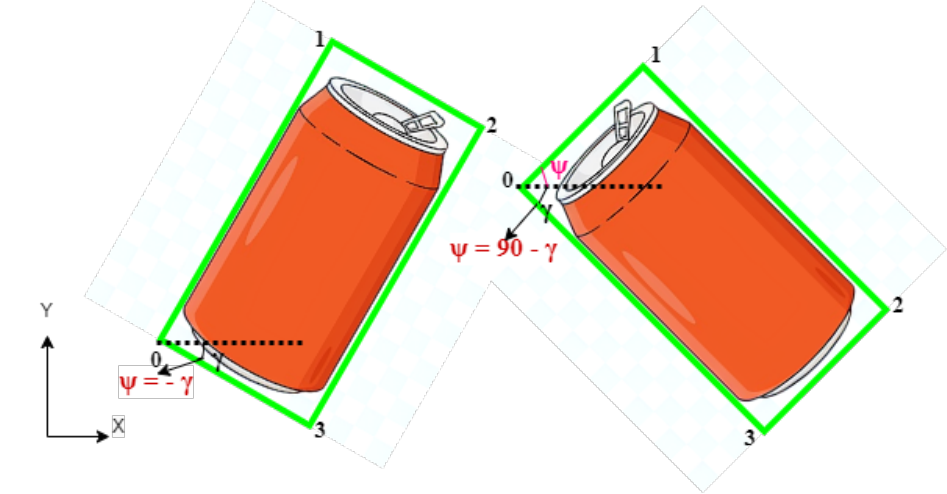}}
\caption{MinAreaRect rotation angle($\gamma$) algorithm explained and converting it into actual rotation angle($\psi$).}
\label{angle}
\end{figure}

The FastSAM algorithm produces masks that highlight each object in an image. These masks can be converted into outlines, allowing to fit a rotated rectangle around each object. This rectangle is defined by four corners, numbered 0 to 3 in a clockwise direction, with the first corner having the smallest $x$-value. The algorithm also provides an "angle" output, representing the tilt of the rectangle. This angle is calculated by measuring the rotation from the $x$-axis. In simpler terms, it helps understand how each object is oriented in the image. Figure~\ref{angle} visually illustrates this process. The "calculated angle"($\gamma$) is determined by the algorithm, while the "actual angle"($\psi$) is the resulting output. The MinAreaRect function assigns an angle within the -90 to 90 degrees range. To enhance clarity,it gets ascertained whether the rectangle's edge between corners 0 and 3 (denoted as $A$) corresponds to its height or width. This determination ensures an accurate representation of the rectangle's rotation angle, providing valuable insights into the object's orientation:
\[
    \begin{cases}
      \text {if} \, A = \text{rectangle's width} \rightarrow \psi = 90 - \gamma\\
      \text {if} \, A = \text{rectangle's\,height} \rightarrow \psi = - \gamma \\
    \end{cases}
\]

The algorithm described above calculates the angle between the $x$-axis and the shorter edge of the object. It is important to note that these angle calculations do not necessitate high precision. The gripping force exerted by the gripper during grasping is designed to accommodate variations in object orientation within its flange. Even if the calculated angle deviates slightly from the real value by a few degrees, the gripper's force ensures effective grasping and manipulation of the objects.

\begin{figure}[ttbp]
\centerline{\includegraphics[width=0.35\textwidth]{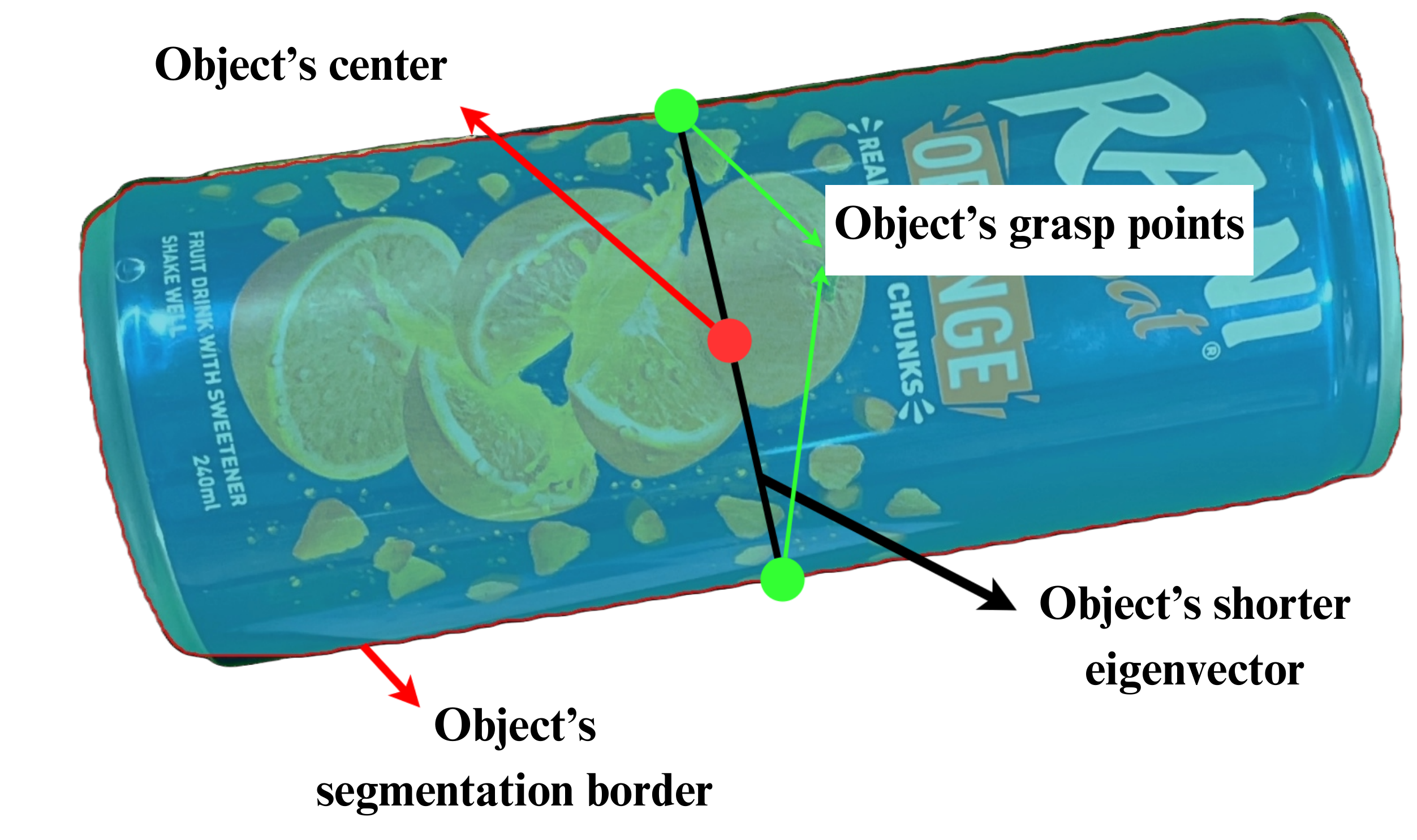}}
\caption{Propsed geometrical method for obtaining grasp points using object's eigenvector.}
\label{abmive}
\end{figure}

\begin{figure}[ttbp]
\centerline{\includegraphics[width=0.5\textwidth]{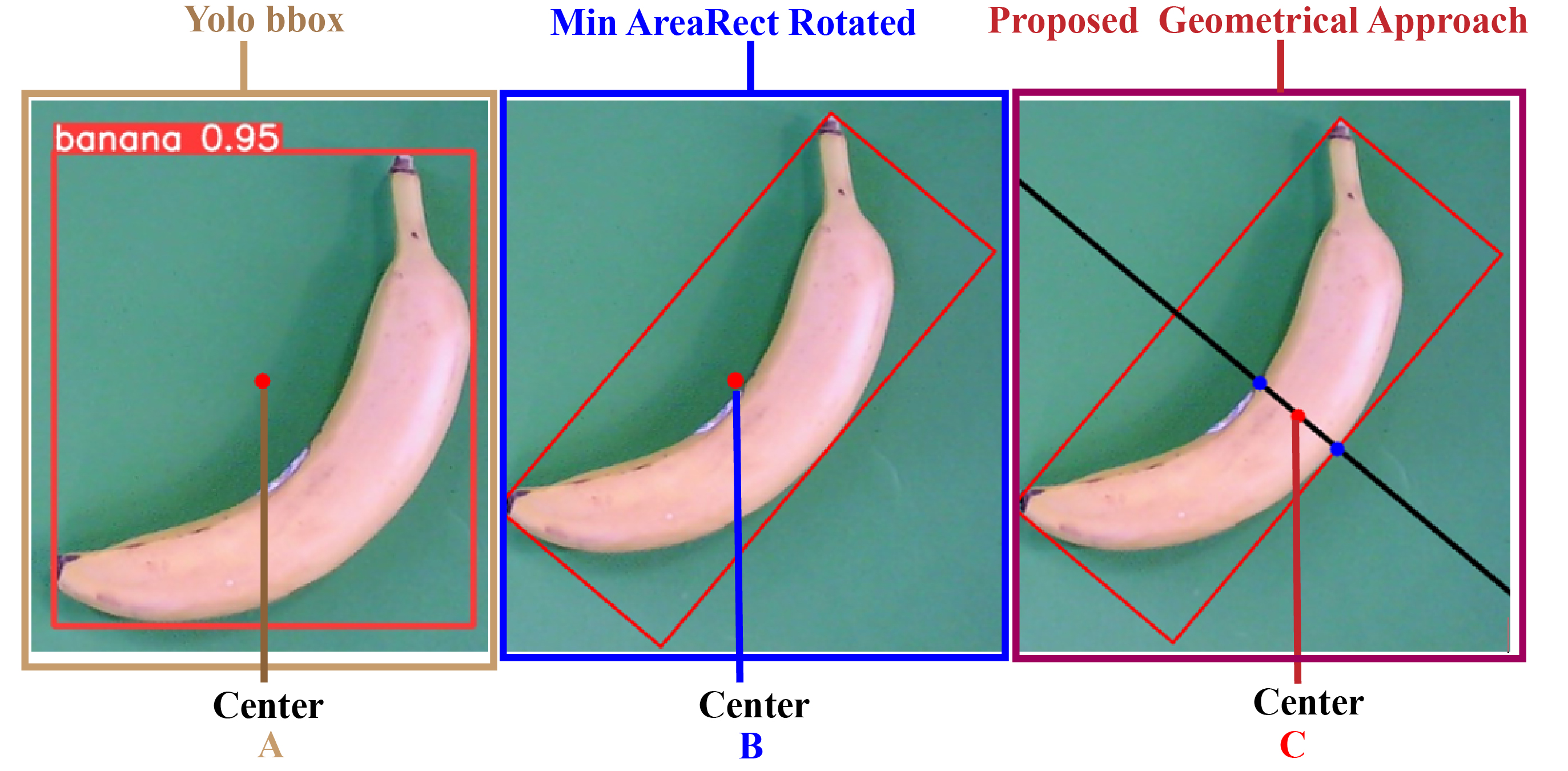}}
\caption{Bounding box and it is center a) output center of YOLO's bounding box b) output center the MinAreaRect function c) output center of proposed geometrical approach.}
\label{moz}
\end{figure}

Using the rectangle obtained in the previous step, the center of the rotated rectangle can be easily found. By considering the object's boundary polygon and a line that crosses the center of the rotated rectangle at its rotation angle, two points that are likely to be the best places to grasp the object can be identified. This line represents the shorter side of the object, crossing its center. The point where this line intersects with the object's boundary polygon gives the solution that is needed. Figure~\ref{abmive} shows a visual representation of this geometric method. The two points that is obtained from this process become the grasp points, and finding the midpoint between them gives the center of the object.
It is worth noting that when dealing with curved or rotated objects, the center of a simple bounding box (a straight rectangle) might not match the actual center of the object. And although the rotated rectangle from the MinAreaRect function might not perfectly indicate the object's center, the solution explained above consistently gives an accurate measurement of the object's center. Figure~\ref{moz} compares these three considerations on a curved object.

Since the DPR lands at the object's center and the gripper rotates with the object, grasp points do not need to be measured as separate coordinates. Only the distance between them is needed , which accurately measures the width of each object for the grasping process. This makes the process simpler while ensuring precise object manipulation.
\subsubsection{Final complete model}
In summary of the model outlined in the preceding sections, each robot grasping attempt follows a systematic sequence. Initially, the robot positions itself at the object's center coordinates on the $x$ and $y$ axes. Subsequently, the rotary component of the gripper aligns itself with the object's rotation angle. The robot then moves along the $z$-axis based on the object's label and the predetermined nominal depth. Concurrently, the gripper opens to the width of the object and endeavors to secure a grasp. Ultimately, the DPR moves to the destination point, the gripper adjusts its opening as required, and the object is successfully placed at the designated destination. This cohesive process ensures efficient and accurate object manipulation by the robot.
\subsection{Challenges and important notes }
\label{sec:challenge}
\subsubsection{Standing objects}
A noteworthy characteristic of many objects in the CPO dataset is their propensity to rest on one side when positioned beneath the robot. However, certain items, such as cans, juices, and cups, possess standing configurations. Contrary to expectations, for efficient packing, these objects need not maintain an upright position and be able to be placed on their sides. Consequently, a distinct labeling system is introduced during annotation to differentiate between objects in standing and lying positions—such as cups labeled based on orientation. This distinction becomes crucial in the subsequent grasping process, as elucidated in section \ref{sec:section6}

\subsubsection{Detecting standing cups}
\begin{figure}[ttbp]
\centerline{\includegraphics[width=0.3\textwidth,angle=90,origin=c]{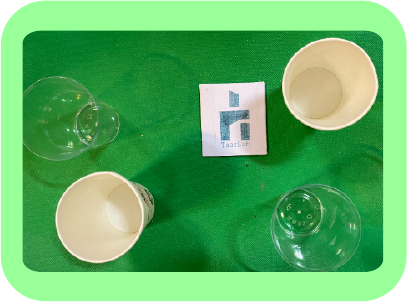}}
\caption{Different views of standing cups in one image,a challenge on detecting these type of objects.}
\label{standing}
\end{figure}

During the object-detecting process(Section \ref{sec:sectionod}), a challenge has been raised for detecting standing cups. Upon testing many possible solutions, it has been inferred that the position of standing cups, with respect to the photo, would make a huge difference in how the camera sees them. Fig.~\ref{standing} illustrates the impact of object placement on the outline shape of the object, making it hard for the trained network to recognize them. Thus, more attention should be paid to adding pictures of different angles and positions on standing cups to the CPO dataset.
\subsubsection{Detecting transparent plastic disposables}
Objects with transparent materials, like clear plastics, pose a challenge for the model due to potential confusion with the background, leading to inaccuracies in the trained model's outcomes. Additionally, distinguishing between forks and spoons, which share a similar shape, presents another potential source of error. While YOLO demonstrates robust object detection capabilities, it encounters difficulties with small objects (those occupying 1/10 of the picture size) \cite{gao2023dms}. However, given that none of the objects in the CPO dataset are considered small objects, this issue has been addressed by augmenting the dataset with sample instances of these labels.

To specifically address the challenges posed by transparent objects and the similarity between forks and spoons,images containing these objects exclusively were introduced. This deliberate addition aids the model in reducing confusion during the training process, enhancing its ability to accurately detect and distinguish these objects within the CPO dataset.

\subsection{The DPR robot and model's data transfered to it}
\begin{figure}[ttbp]
\centerline{\includegraphics[width=0.33\textwidth]{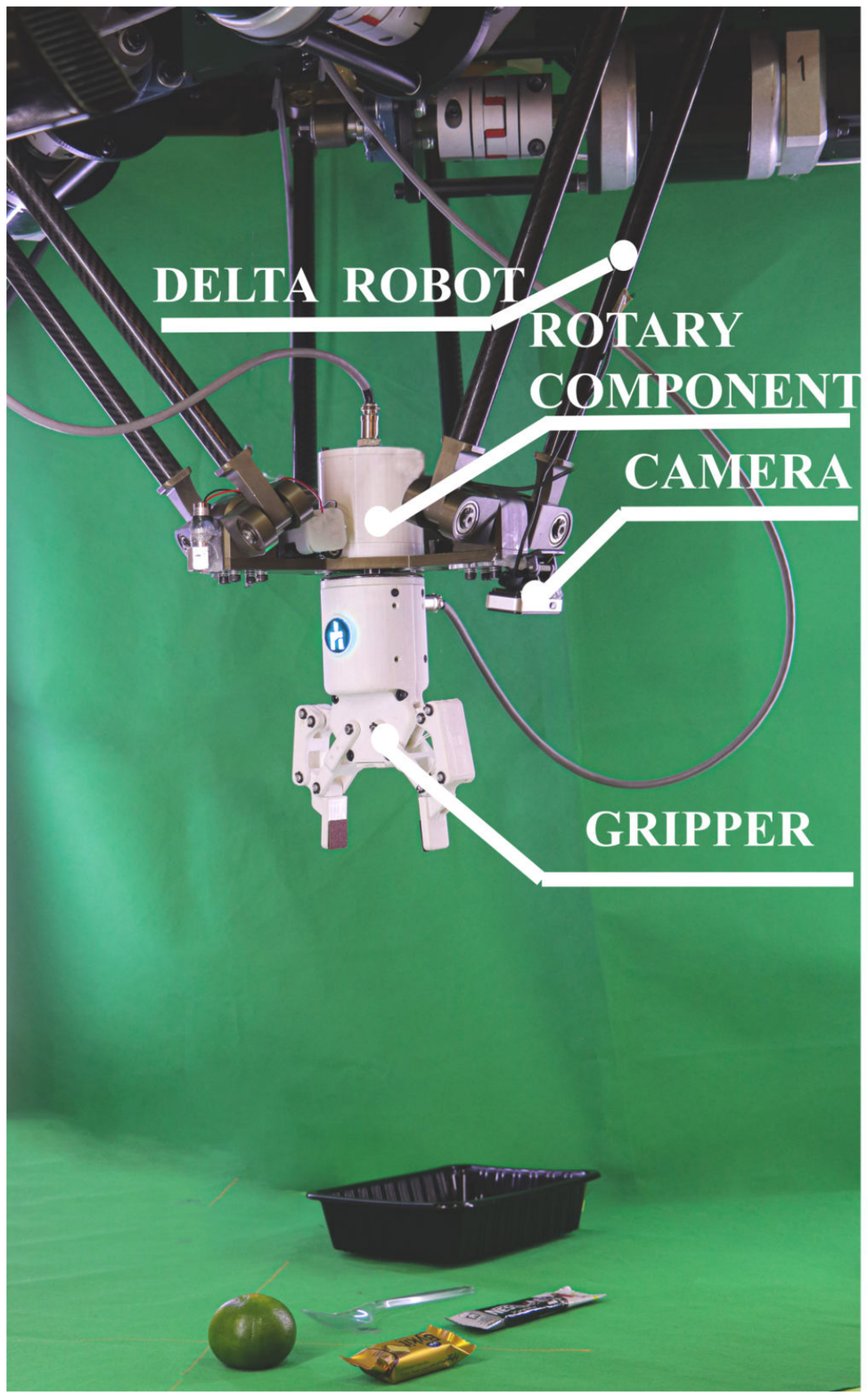}}
\caption{The experimental environment of this study which includes 3-DOF DPR, the gipper and the objects.}
\label{gripper}
\end{figure}

As outlined in the introduction, the Delta parallel robot (DPR) boasts three translational degrees of freedom (DOF), rendering it well-suited for pick-and-place applications\cite{abed2022dynamics}. Fig.\ref{gripper} illustrates the incorporation of a gripper at the End-Effector (EE) of the DPR, inspired by the 2F88 gripper developed by Robotiq \cite{allen2013gripper}. This gripper is exclusively crafted for research purposes and no commercial development intentions. To cater to packaging needs requiring one rotational DOF about the axis perpendicular to the EE, a rotary component is affixed to the gripper, as depicted in Fig.\ref{gripper}. Consequently, the entire robot possesses the capability to execute three translational motions in $(x, y, z)$ and one rotational DOF, specifically the yaw angle, constituting Schönflies motion\cite{tamizi2022experimental}. The gripper excels at grasping objects enclosed within a circle with a diameter of less than 8 cm. Each flange of the gripper is equipped with a force sensor, providing feedback on the force applied by the fingers on the grasped object. In this study, this sensor serves the purpose of determining whether an object is grasped or released \cite{yarmohammadi2023experimental}. Figure~\ref{gripper} showcases the gripper's integration with the DPR. The algorithm's output data is transmitted wirelessly to the robot using the Transmission Control Protocol (TCP), with a connection established to the gripper via a data cable.
However, it is crucial to note that all calculations are initially performed in pixels. To ensure accurate coordination with the robot, these pixel-based measurements need to be converted to the robot's exact coordinates. Achieving this necessitates a precise calibration of the robot. The chosen solution involves capturing images from multiple heights using the robot's webcam. By marking the positions of corners on a gridded paper beneath the robot, a transformation and offset matrix is determined. This matrix facilitates the conversion of camera-based measurements to the robot's coordinate system through the application of the robot's forward kinematics\cite{pasiar2023generating}.
\section{Experimental test studies and results}
\label{sec:section6}
Table~\ref{grasp} summarizes the real-time experimental study on grasping various classes, presenting results in two key aspects: object detection and grasping. It is essential to note that even if the proposed model misclassifies some objects, successful grasping is still achieved due to accurate attribute calculations. Conversely, certain objects, such as teabags, may exhibit good detection results, but their challenging geometrical shapes, particularly their depth, make grasping nearly impossible for the gripper. In contrast, objects like Nescafe packs, spoons, forks, and knives, despite their lack of depth, are successfully grasped by the gripper.

It is important to highlight the challenge posed by grasping standing objects, as discussed in Section \ref{sec:challenge}. The robot intentionally displaces these objects (cups, juices, and cans in a standing state), causing them to fall. Subsequently, the robot treats them as if they were in a laying position and attempts to grasp them. 

For each pack manipulated by the DPR, a consistent three-step process is executed:

1)The user places a pre-assembled pack at a designated location under the robot. The DPR then positions itself above the pack and captures an image.

2)Subsequently, the DPR returns to its "home" point, where it has visibility of objects on the ground. The webcam captures an image of this scene.

3)Both images undergo processing by the network outlined in Fig.~\ref{koll}. The resulting output from this network provides essential information for each object, including coordinates and rotation angle for both the origin and destination, as well as the width and label of the object.

\begin{figure}[ttbp]
\centerline{\includegraphics[width=0.5\textwidth]{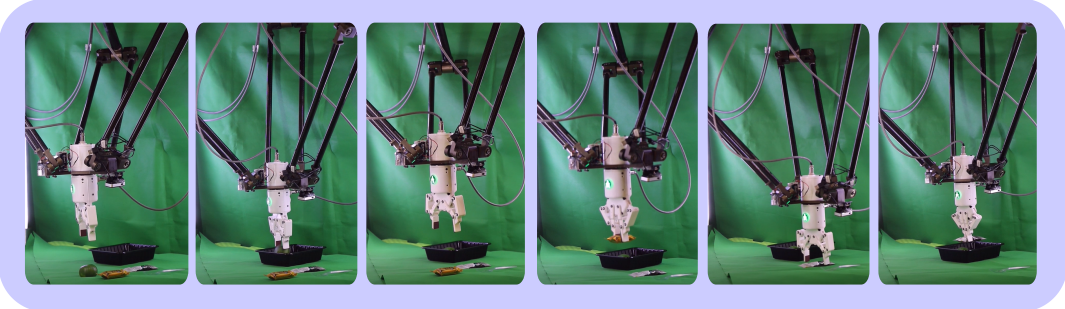}}
\caption{Sequences of grasping the food packages using the DPR.}
\label{grasping}
\end{figure}

The DPR, in conjunction with the gripper, proceeds to process the acquired data. In a carefully orchestrated sequence, the robot adeptly manipulates all the necessary elements except for the pack itself—ensuring the pack remains stationary. Through the meticulously designed process, the robot successfully recreates each custom pack with precision and efficiency. Fig.~\ref{grasping} shows a few steps of grasping the objects and putting them in the packages.\footnote{\href{https://drive.google.com/file/d/1t8XASfcEXB5aQR1wbsKf4aBEIp\_OQXfN/view?usp=sharing}{Supplementary material}}.
\begin{table}[ttbp]
\caption{Pyhsical grasping comparison.}
\begin{center}
\resizebox{8cm}{!}{
\begin{tabular}{c|c|c}
\textbf{label}&\textbf{detected} & \textbf{\textit{physical grasping}}\\
Banana&10/10&10/10 \\
\cline{1-3} 
Biscuit&9/10&6/10  \\
\cline{1-3} 
Cake&10/10&9/10  \\
Cup(laying)&10/10&10/10 \\
\cline{1-3} 
Cup(standing)&7/10&8/10\\
\cline{1-3} 
Fork&8/10&10/10\\
\cline{1-3} 
Juice(laying)&9/10&10/10\\
\cline{1-3} 
Juice(standing)&9/10&7/10 \\
\cline{1-3} 
Knife&9/10&5/10\\
\cline{1-3} 
Nescafe&9/10&10/10\\
\cline{1-3} 
Nescafe(square)&8/10&8/10\\
\cline{1-3} 
Pack&8/10& -\\
\cline{1-3} 
Rani(laying)&10/10&10/10\\
\cline{1-3} 
Rani(standing)&9/10&8/10\\
\cline{1-3} 
Spoon&9/10&10/10\\
\cline{1-3} 
Straw&10/10&8/10\\
\cline{1-3} 
Tangerine&10/10&10/10\\
\cline{1-3} 
Teabag&7/10&0/10\\
\cline{1-3} 
Average(\%)&89.4&81.7\\
\end{tabular}
}
\label{grasp}
\end{center}
\end{table}

\section{Conclusion}
\label{sec:section7}
This study explored automating catering package packing with a 3-degree-of-freedom DPR, focusing on object detection using deep learning. A dataset of 1500 images with over 4000 labels across 19 classes was established, achieving 93\% accuracy with YOLOV5. Geometric methods determined object properties and grasp points. FastSAM segmentation enhanced grasp point extraction, making the model versatile. Data transmission to the robot facilitated seamless integration. Experimental results showed an 85\% success rate in autonomously assembling packages. This marks a significant stride toward enhancing automation in the food industry's packaging processes, affirming the feasibility and efficiency of employing a Delta parallel robot in such applications. As part of the ongoing research for this study, the exploration will extend to employing graph neural networks  and also using newer object detection tools like YoloV9, a more intricate and advanced approach compared to existing models. This strategic shift aims to introduce a heightened level of complexity to the detection process, pushing the boundaries of current methodologies and paving the way for enhanced capabilities in automated packaging systems.

\bibliographystyle{IEEEtran}
\bibliography{conference_101719.bib}
\end{document}